\documentclass[twocolumn]{article} 
\usepackage{multicol} 
\usepackage{amsmath}        
\usepackage{lipsum}         
\usepackage{booktabs}       
\usepackage{stfloats}       
\usepackage{algorithm}      
\usepackage{algorithmic}    
\usepackage{tikz}           
\usetikzlibrary{shapes,arrows}

\tikzstyle{block} = [rectangle, draw, text width=3cm, text centered, rounded corners, minimum height=4em]
\tikzstyle{input} = [coordinate]
\tikzstyle{output} = [coordinate]
\tikzstyle{line} = [draw, -latex']

\usepackage[utf8]{inputenc} 
\usepackage[T1]{fontenc}    
\usepackage{graphicx}       
\usepackage{url}            
\usepackage{hyperref}       
\usepackage[numbers,sort&compress]{natbib} 
\usepackage{geometry}       
\title{Deep Learning-Based BMD Estimation from Radiographs with Conformal Uncertainty Quantification}
\author{Long Hui\thanks{Sky Long Artificial Intelligence Company. Email: longhui@berkeley.edu} \and Wai Lok Yeung\thanks{Corresponding author. Email: wailokyeung@institution.edu}}
\date{\today} 

\begin{document}

\twocolumn[
\begin{@twocolumnfalse}
\maketitle
\begin{abstract}
Osteoporosis screening remains a critical yet underutilized healthcare intervention due to limited access to gold-standard Dual-energy X-ray Absorptiometry (DXA). This paper presents a proof-of-concept methodology for using widely available plain radiographs in opportunistic Bone Mineral Density (BMD) estimation through deep learning, with particular focus on robust uncertainty quantification—a crucial requirement for future clinical applications. We develop an EfficientNet-based model to predict BMD from bilateral knee radiographs using the Osteoarthritis Initiative (OAI) dataset, and implement two distinct Test-Time Augmentation (TTA) approaches to enhance prediction stability: traditional averaging-before-conformalization and a multi-sample approach treating each augmented version independently. Most importantly, we apply Split Conformal Prediction to generate statistically rigorous patient-specific prediction intervals with guaranteed coverage properties. Our methodology achieves a Pearson correlation of 0.68 with traditional TTA, while providing reliable uncertainty bounds at various confidence levels (90\%, 95\%, 99\%). We demonstrate that while the traditional TTA approach yields better point prediction performance, the multi-sample approach can produce slightly tighter confidence intervals while maintaining proper coverage. The conformal prediction framework appropriately expresses higher uncertainty for more challenging cases, demonstrating that the model effectively "knows when it doesn't know." While the current results present limitations for immediate clinical deployment due to anatomical site mismatch between knee radiographs and standard DXA measurements, our approach establishes a methodological foundation for trustworthy AI-assisted BMD screening that could guide future research towards improving early detection of osteoporosis using already acquired radiographs in routine clinical practice.
\end{abstract}
\vspace{0.5cm} 
\end{@twocolumnfalse}
]

\section{Introduction}
\label{sec:intro}

The early detection and treatment of osteoporosis, a systemic skeletal disease characterized by decreased bone mass and increased fracture risk, remains a critical yet underutilized healthcare intervention. Approximately half of women and one-fifth of men over 50 will experience an osteoporotic fracture in their remaining lifetime \cite{mccloskey2021iof}, with substantial associated morbidity, mortality, and healthcare costs \cite{hernlund2013IOF}. Despite these serious consequences, osteoporosis screening rates remain suboptimal globally \cite{chotiyarnwong2020global}. 

Dual-energy X-ray Absorptiometry (DXA) is the gold standard for measuring Bone Mineral Density (BMD) and diagnosing osteoporosis \cite{who_criteria}. However, limited access to DXA scanners, particularly in resource-constrained settings, impedes widespread screening. Plain radiographs, in contrast, are widely available and routinely obtained for various clinical indications. If these could be leveraged for opportunistic BMD assessment, screening coverage could potentially be expanded significantly and at minimal additional cost.

Deep learning has shown promise in analyzing medical images to extract information that may not be apparent to human observers \cite{esteva2017nature, gulshan2016jama}. Several studies have previously investigated the use of deep learning to predict BMD from conventional radiographs \cite{yamamoto2020scientific, thian2021knees, liu2023analysis, thian2022hand, pinnamaneni2021use, lindsey2023deep, wang2023hand}. However, for such models to be clinically useful, they must not only provide accurate predictions but also reliable estimates of prediction uncertainty—a requirement for responsible clinical decision-making.

Conformal prediction offers a distribution-free framework for constructing prediction intervals with guaranteed coverage, making it particularly suitable for medical applications \cite{Angelopoulos2021GentleIntroduction, kompa2021second}. When combined with deep learning, it provides a principled approach to uncertainty quantification that avoids the need for extensive model recalibration or explicit modeling of the prediction distribution.

Bone Mineral Density (BMD) is a cornerstone in the assessment of bone health, playing a pivotal role in diagnosing osteoporosis and predicting fracture risk. While Dual-energy X-ray Absorptiometry (DXA) remains the gold standard for BMD measurement, its accessibility can be limited by cost and availability, particularly in resource-constrained settings or for widespread population screening. Furthermore, DXA involves a dedicated procedure and radiation exposure, making it less ideal for opportunistic screening. Plain film radiography (X-rays), on the other hand, are among the most common medical imaging procedures globally, performed for a myriad of clinical indications. The ubiquity and low cost of X-rays present a significant, yet largely untapped, opportunity for developing accessible, opportunistic BMD screening tools. Such tools could identify individuals at risk of low BMD who might not otherwise undergo dedicated testing, facilitating earlier intervention and potentially reducing fracture incidence.

Large-scale, longitudinal studies such as the Osteoarthritis Initiative (OAI) have provided invaluable, publicly available datasets that include knee radiographs and associated clinical data, including BMD measures for subsets of participants. Such resources have been instrumental in advancing research into image-based biomarkers and AI applications in musculoskeletal conditions.

Artificial intelligence (AI), particularly deep learning (DL), has demonstrated remarkable potential in extracting quantitative information from standard medical images, including radiographs. This capability has spurred research into using DL models for BMD estimation or osteoporosis classification, leveraging datasets like OAI for knee studies or other similar cohorts for various anatomical sites \cite{Yan2021arXiv, Sato2022Biomedicines, Golestan2023JMAI, Hsieh2021NatCommun, Jang2021SciRep, Yamamoto2020Biomolecules}. However, a critical barrier to the clinical adoption of these AI tools is the need for robust uncertainty quantification (UQ). For AI-driven diagnostic aids to be trustworthy, they must not only provide accurate predictions but also reliably communicate their confidence, as "black box" predictions without such context can be misleading \cite{kompa2021second}.

To address this imperative for trustworthy AI in medical imaging, we investigate the application of Conformal Prediction. This framework is model-agnostic, distribution-free, and capable of generating statistically rigorous, patient-specific prediction intervals for BMD estimates. In this study, we develop an EfficientNet-based deep learning model for BMD regression using bilateral knee radiographs from the Osteoarthritis Initiative (OAI) dataset. We then apply Conformal Prediction to the model's outputs, potentially enhanced by Test-Time Augmentation (TTA). 

It is important to note a fundamental limitation of this approach: we predict BMD values derived from DXA measurements of the femoral neck using knee radiographs. While both measurements relate to bone health, the anatomical site mismatch introduces inherent challenges in establishing direct correlations. This work should therefore be viewed as an exploration of methodological approaches for uncertainty quantification in BMD estimation, rather than a clinically ready diagnostic tool. The primary contribution is the demonstration and evaluation of this methodology for providing dependable uncertainty bounds for BMD estimates, a vital step towards building more reliable AI tools for opportunistic osteoporosis screening.

The remainder of this paper is organized as follows: Section \ref{sec:related_work} reviews related work in AI-based BMD estimation and uncertainty quantification in medical imaging. Section \ref{sec:methods} details our bilateral knee dataset, the BMD regression model, training procedures, Test-Time Augmentation strategy, and the Conformal Prediction methodology. Section \ref{sec:results} presents the performance of our regression model and the characteristics of the generated conformal prediction intervals. Section \ref{sec:discussion} discusses the implications of our findings, the importance of uncertainty quantification, limitations of the current study, and directions for future research. Finally, Section \ref{sec:conclusion} concludes the paper.

\section{Related Work}
\label{sec:related_work}
The application of artificial intelligence, particularly deep learning, to medical imaging has shown considerable promise for various diagnostic and screening tasks. In the context of bone health, several studies have explored the use of deep learning models to estimate Bone Mineral Density (BMD) or screen for osteoporosis using plain radiographs, which are more widely accessible than specialized DXA scans. For instance, researchers have developed algorithms to predict osteoporosis from hip radiographs \cite{Jang2021SciRep, Yamamoto2020Biomolecules}, chest radiographs \cite{Yan2021arXiv, Sato2022Biomedicines}, and dental panoramic radiographs \cite{Gaudin2024Diagnostics, yamamoto2021development}. These efforts demonstrate the potential of AI to enable opportunistic screening and improve early detection of low BMD, though the reliability and generalizability of these predictions remain active areas of research. Our work builds upon this foundation by not only predicting BMD but also by providing a rigorous measure of uncertainty for these predictions.

Recently, numerous studies have explored the use of deep learning for osteoporosis screening from radiographs of various anatomical sites, including chest X-rays \cite{Yan2021arXiv, Sato2022Biomedicines}, hand and wrist X-rays \cite{Kim2023arXiv, Golestan2023JMAI}, spine X-rays \cite{Mao2022FrontEndocrinol, Zhang2020Bone, Lee2020SkeletalRadiol, wang2021deep}, and hip X-rays \cite{Hsieh2021NatCommun, Jang2021SciRep, Yamamoto2020Biomolecules}. A recent systematic review by Liu et al. \cite{PMC11117497, Liu2024EJRadiol} highlights the growing body of evidence supporting the efficacy of AI methods for osteoporosis classification and fracture risk assessment using radiographs.

While AI models can achieve high predictive accuracy, their adoption in critical clinical settings is often hampered by their "black-box" nature and the lack of reliable uncertainty quantification (UQ). In medical applications, understanding the confidence of a model's prediction is paramount for responsible decision-making, risk assessment, and building trust with clinicians \cite{kompa2021second, He2024JIMR}. An erroneous prediction with high confidence can have severe consequences, whereas a prediction accompanied by an appropriate level of uncertainty can prompt further investigation or caution. Various UQ methods have been proposed, including Bayesian neural networks, ensemble methods, and quantile regression, each with its own set of assumptions and computational trade-offs.

Conformal Prediction (CP) has emerged as a powerful and versatile framework for UQ that offers distribution-free coverage guarantees, meaning that the generated prediction intervals are valid under minimal assumptions about the underlying data distribution or model \cite{vovk2005algorithmic, shafer2008tutorial}. This is particularly advantageous in complex and safety-critical medical domains where data distributions are often unknown, difficult to model, or where models may encounter out-of-distribution samples for which they might otherwise produce confident but incorrect predictions \cite{Angelopoulos2021GentleIntroduction}. CP can be applied to any pre-trained model, making it a practical add-on for existing AI systems. In medical imaging, CP has been utilized for various tasks, including classification and segmentation, to provide more reliable outputs. For regression tasks, such as the BMD estimation in this paper, methods like Conformalized Quantile Regression (CQR) and its variants have been developed to produce prediction intervals with guaranteed marginal coverage \cite{romano2019conformalized}. These intervals help in understanding the range within which the true BMD value is likely to fall, providing a transparent and reliable way to communicate model uncertainty, which is crucial for clinical interpretation and enabling clinicians to make more informed decisions. Our work leverages CP to provide such rigorous uncertainty bounds for BMD estimates derived from combined hand and knee radiographs.
\section{Methods}
\label{sec:methods}

\subsection{Theoretical Background and Formulation}
\label{sec:methods_theory}

\subsubsection{DXA and BMD Diagnostic Criteria}
\label{sec:methods_theory_dxa}
Dual-energy X-ray Absorptiometry (DXA) is the current clinical standard for measuring Bone Mineral Density (BMD). Results are typically reported as T-scores, which represent the number of standard deviations by which a patient's BMD differs from the mean BMD of a young, healthy adult reference population of the same sex. According to the World Health Organization (WHO) criteria, osteoporosis is diagnosed at a T-score of -2.5 or lower, osteopenia (low bone mass) is defined by a T-score between -1.0 and -2.5, and normal BMD is a T-score of -1.0 or higher. These classifications are crucial for assessing fracture risk and guiding treatment decisions.

\subsubsection{BMD Regression Task}
\label{sec:methods_theory_regression}
Our primary task is to estimate BMD from plain radiographs, which we frame as a regression problem. Given an input radiograph $X_i$, the goal is to predict its corresponding continuous BMD value $Y_i$. We train a deep learning model $f_\theta$ with parameters $\theta$ to minimize a chosen loss function over a training dataset $\{(X_i, Y_i)\}_{i=1}^{N_{train}}$. While Mean Squared Error (MSE) is common for regression, it can be sensitive to outliers. The Huber Loss offers a compromise between MSE and Mean Absolute Error (MAE), being quadratic for small errors and linear for large errors, thus providing more robustness to outliers. For our model, the Huber Loss with parameter $\delta=0.5$ was used, which applies quadratic loss for absolute errors smaller than 0.5 and linear loss for larger errors. This helps the model handle potential outliers in the BMD measurement data while still maintaining sensitivity to smaller prediction errors.

\subsubsection{Test-Time Augmentation (TTA)}
\label{sec:methods_theory_tta}
Test-Time Augmentation (TTA) is a technique used to improve the robustness and predictive performance of machine learning models, particularly in computer vision. Instead of making a prediction based on a single instance of the input, TTA involves creating multiple augmented versions of each test sample (e.g., through flips, rotations, crops) and then aggregating the model's predictions across these versions (e.g., by averaging). This process can help reduce the impact of small variations in the input that might otherwise lead to unstable predictions, effectively providing a more ensembled prediction from a single trained model.

\subsubsection{Split Conformal Prediction}
\label{sec:methods_theory_cp}
Conformal Prediction (CP) provides a framework for generating statistically valid prediction intervals for the outputs of any machine learning model, including regression models \cite{vovk2005algorithmic, shafer2008tutorial}. In the common split-conformal approach, the available data is divided into a training set, used to fit the model $f_\theta$, and a separate calibration set $\{(X_j, Y_j)\}_{j=1}^{N_{\text{calib}}}$. Non-conformity scores, $s_j = \mathcal{A}(f_\theta(X_j), Y_j)$, are computed for each sample in the calibration set. These scores measure how "unusual" or "non-conforming" each calibration sample is with respect to the model's prediction; a common choice for regression is the absolute residual, $s_j = |Y_j - f_\theta(X_j)|$. For a new test sample $X_{test}$ and a desired significance level $\alpha$ (e.g., $\alpha = 0.1$ for 90\% prediction intervals), the $(1-\alpha)(N_{\text{calib}}+1)$-th smallest non-conformity score from the calibration set, denoted $q_{1-\alpha}$, is used to form the prediction interval. For regression, the interval is typically constructed as $[f_\theta(X_{test}) - q_{1-\alpha}, f_\theta(X_{test}) + q_{1-\alpha}]$. This procedure guarantees that, on average, the true value $Y_{test}$ will be contained within the predicted interval with a probability of at least $1-\alpha$, under the assumption that the calibration and test data are exchangeable (e.g., drawn i.i.d. from the same distribution).

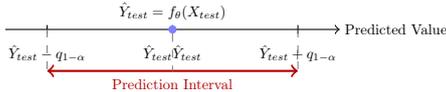
\begin{figure}[!t]
    \centering
    \begin{tikzpicture}[scale=0.55, transform shape]
        \draw[->] (-4,0) -- (4,0) node[right] {Predicted Value};
        
        \foreach \x/\xtext in {-3/-q_{1-\alpha}, 0/\hat{Y}_{test}, 3/+q_{1-\alpha}}
            \draw (\x cm,3pt) -- (\x cm,-3pt) node[below=4pt] {$\hat{Y}_{test} \xtext$};
        
        \node[circle, fill=blue!50, inner sep=2pt, label={above:$\hat{Y}_{test} = f_\theta(X_{test})$}] (pred) at (0,0) {};
        
        \draw[<->, thick, red!70!black] (-3, -1.0) -- (3, -1.0) node[midway, below=2pt, text width=4cm, text centered] {Prediction Interval};
        
        \draw[dashed, gray] (pred) -- (0, -1.0);
        \draw[dashed, gray] (-3,0) -- (-3, -1.0);
        \draw[dashed, gray] (3,0) -- (3, -1.0);
    \end{tikzpicture}
    \caption{Conceptual illustration of a symmetric conformal prediction interval for a regression task. $\hat{Y}_{test}$ is the model's point prediction, and $q_{1-\alpha}$ is derived from the calibration set to ensure at least $1-\alpha$ coverage. The prediction interval $[\hat{Y}_{test} - q_{1-\alpha}, \hat{Y}_{test} + q_{1-\alpha}]$ guarantees that $P(Y_{test} \in \text{Interval}) \geq 1-\alpha$.}
    \label{fig:conceptual_cp_interval}
\end{figure}
\subsection{Experimental Setup}
\label{sec:methods_experiment_setup}

\subsubsection{Dataset and Preprocessing}
\label{sec:methods_dataset_exp}
Our study utilized data derived from the Osteoarthritis Initiative (OAI) public use datasets. We focused on subjects who had both knee radiographs and corresponding DXA-derived neck BMD values. Following a procedure similar to Golestan et al. \cite{Golestan2023JMAI}, we included data pairs where the X-ray image and the DXA scan were performed within 180 days of each other. The dataset was further processed by separating left and right knee images from bilateral radiographs, effectively augmenting the number of available image-BMD pairs. The model was configured to accept input images of size 384x384 pixels.

After filtering and processing, the data was split into training, validation, testing, and a dedicated calibration set for conformal prediction, using approximate ratios of 70\%, 10\%, 10\%, and 10\%, respectively. All radiographic images were resized to a standard input size suitable for the EfficientNet architecture and normalized using mean and standard deviation values typically employed for ImageNet-pre-trained models. We refer to the dataset containing both left and right knee images (from bilateral radiographs) used for training our primary models as the "Bilateral Knee" dataset throughout this paper.

\begin{figure}[!t]
    \centering
    \includegraphics[width=0.95\columnwidth]{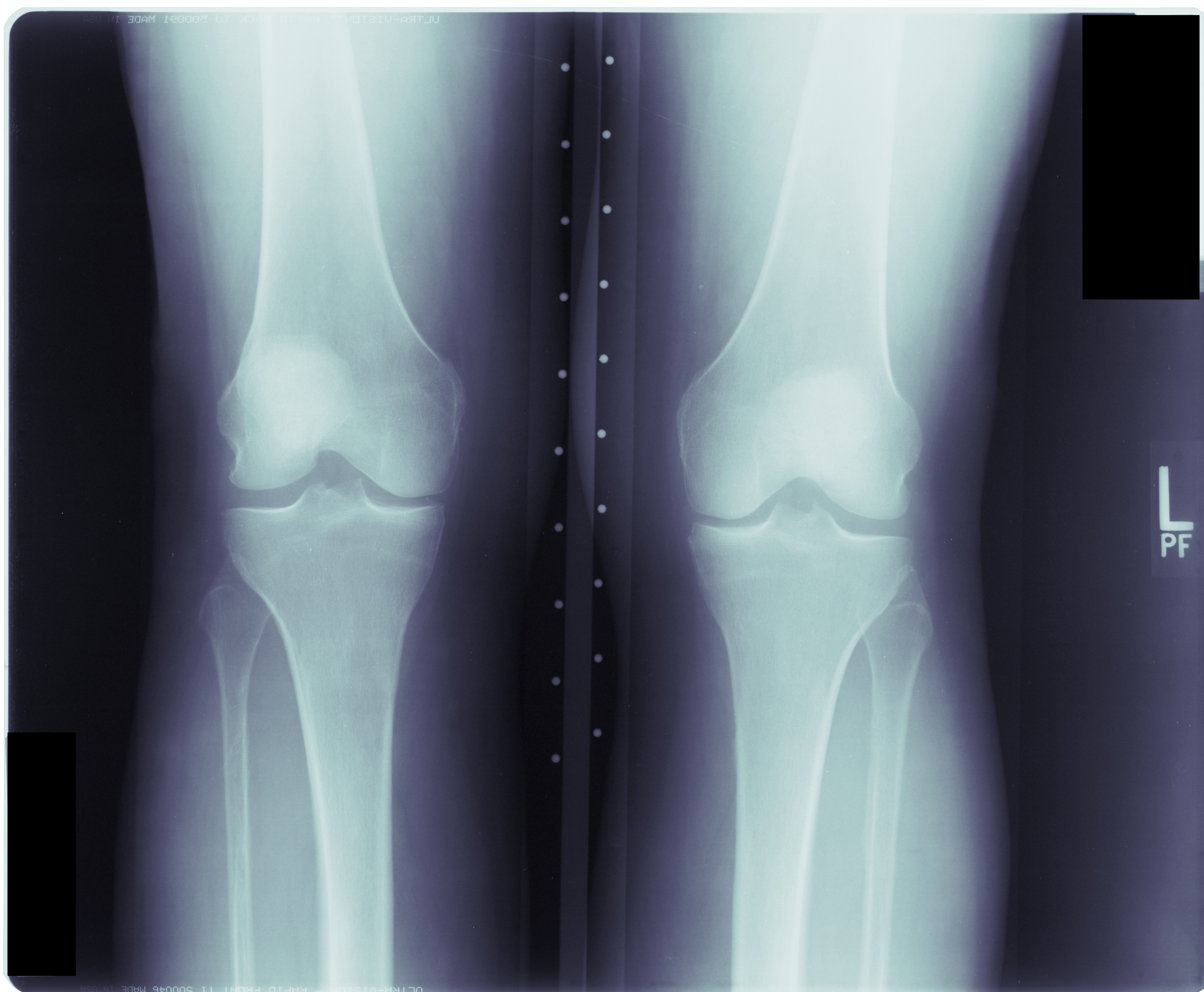}
    \caption{Example of a knee radiograph from the OAI dataset before preprocessing.}
    \label{fig:example_xray_original}
\end{figure}

\begin{figure}[!t]
    \centering
    \begin{tabular}{c@{\hspace{0.5cm}}c}
        \includegraphics[width=0.42\columnwidth]{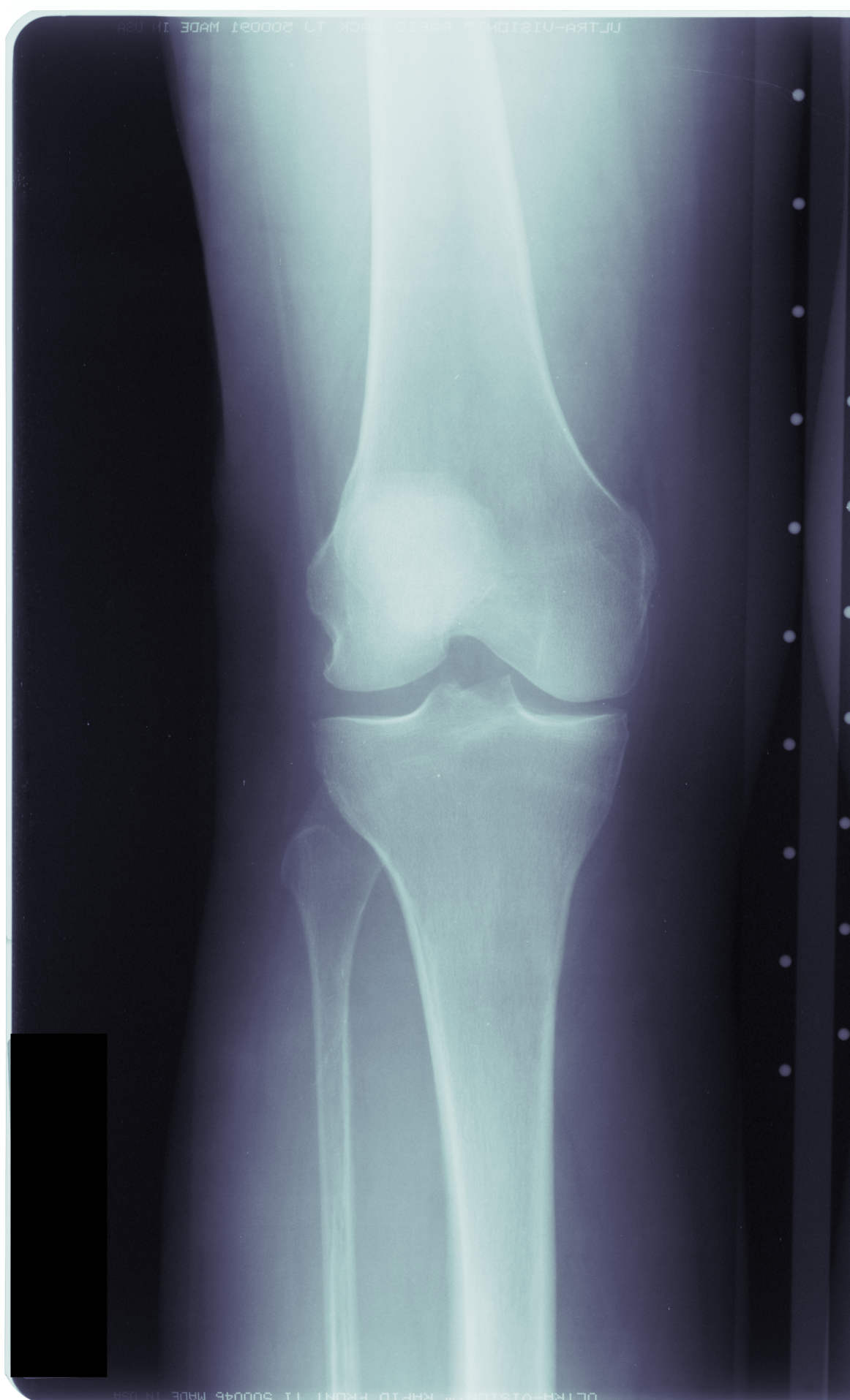} &
        \includegraphics[width=0.42\columnwidth]{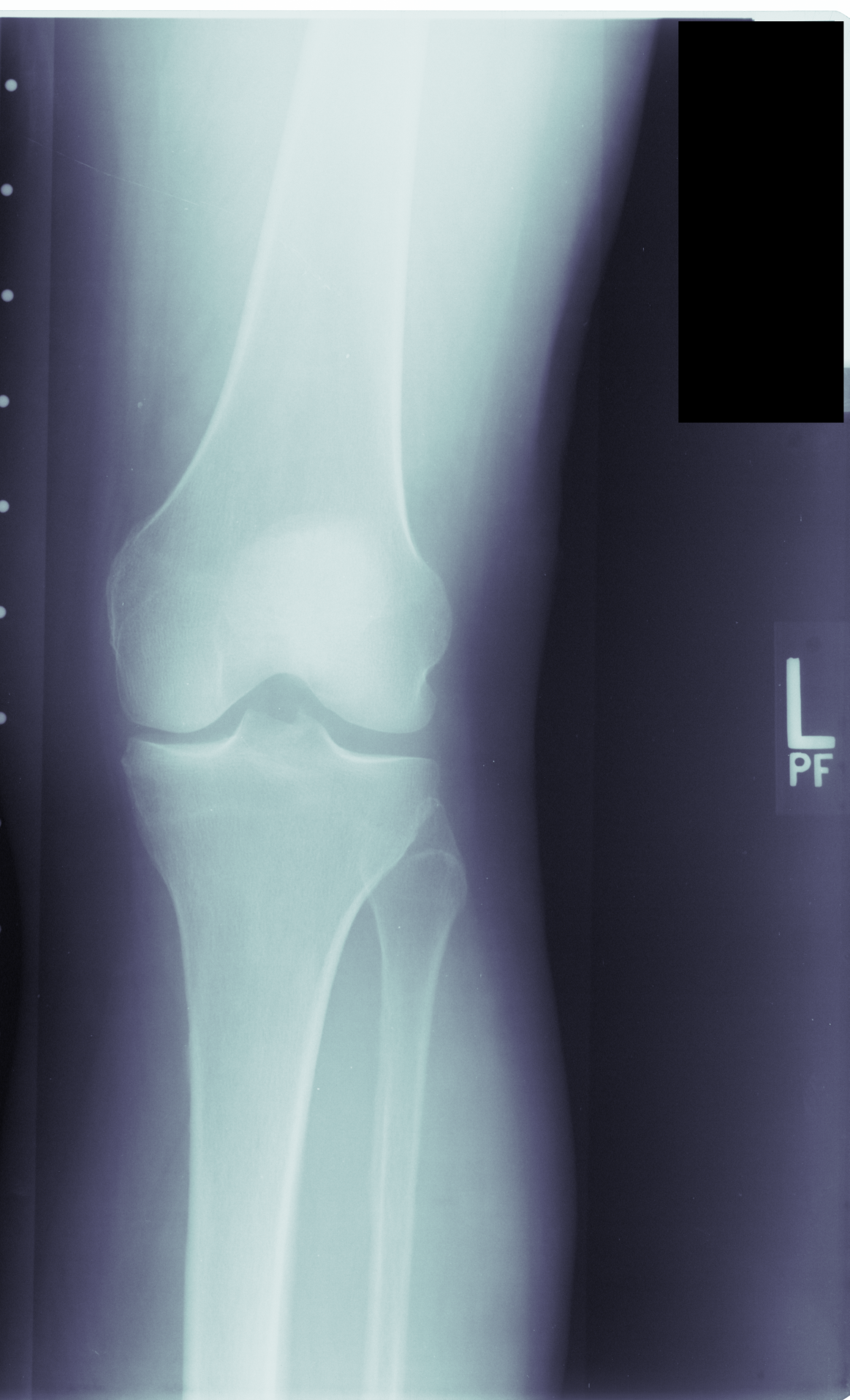} \\
        (a) Left knee & (b) Right knee
    \end{tabular}
    \caption{Example of left and right knee radiographs after splitting the original image during preprocessing.}
    \label{fig:example_xray_split}
\end{figure}

\subsubsection{BMD Regression Model Architecture}
\label{sec:methods_model_exp}
The core of our BMD estimation system is a deep convolutional neural network (CNN) based on the EfficientNet architecture \cite{Tan2019EfficientNet}. Specifically, we employed the \texttt{tf\_efficientnetv2\_m.in21k\_ft\_in1k} model variant, obtained using the \texttt{timm} library and pre-trained on the ImageNet dataset. The pre-trained backbone was augmented with a custom regression head consisting of one or more fully connected layers and a dropout layer with a rate of 0.4 to mitigate overfitting. The final layer outputs a single continuous value representing the predicted BMD.

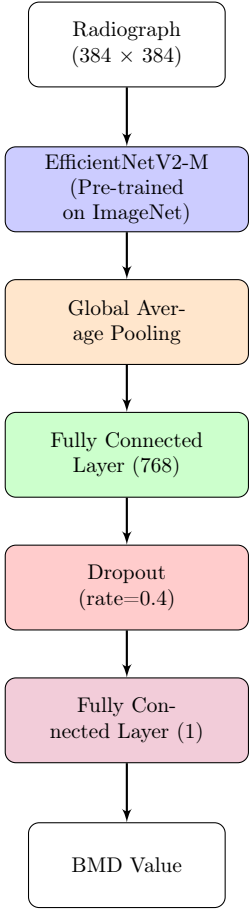
\begin{figure}[!t]
    \centering
    \begin{tikzpicture}[scale=0.8, transform shape, auto, node distance=2cm,>=latex']
        \node [block, fill=blue!20, minimum width=4cm] (efficientnet) {EfficientNetV2-M\\(Pre-trained on ImageNet)};
        \node [block, below of=efficientnet, yshift=-0.2cm, fill=orange!20, minimum width=4cm] (gap) {Global Average Pooling};
        \node [block, below of=gap, yshift=-0.2cm, fill=green!20, minimum width=4cm] (fc1) {Fully Connected Layer (768)};
        \node [block, below of=fc1, yshift=-0.2cm, fill=red!20, minimum width=4cm] (dropout) {Dropout (rate=0.4)};
        \node [block, below of=dropout, yshift=-0.2cm, fill=purple!20, minimum width=4cm] (fc2) {Fully Connected Layer (1)};
        
        \node [block, above of=efficientnet, yshift=0.4cm] (input) {Radiograph \\ (384 × 384)};
        \node [block, below of=fc2, yshift=-0.4cm] (output) {BMD Value};
        
        \draw [->, thick] (input) -- (efficientnet);
        \draw [->, thick] (efficientnet) -- (gap);
        \draw [->, thick] (gap) -- (fc1);
        \draw [->, thick] (fc1) -- (dropout);
        \draw [->, thick] (dropout) -- (fc2);
        \draw [->, thick] (fc2) -- (output);
    \end{tikzpicture}
    \caption{Architecture of the EfficientNetV2-based BMD regression model.}
    \label{fig:model_architecture}
\end{figure}

\subsubsection{Model Training Details}
\label{sec:methods_training_exp}
The BMD regression models were trained using the PyTorch deep learning framework on NVIDIA GPUs. We employed the AdamW optimizer \cite{Loshchilov2017DecoupledWD} with a learning rate of 0.0005 and a weight decay of 0.01 (L2 regularization). A cosine annealing learning rate scheduler with warm restarts was utilized, configured with $T_0=10$, $T_{\text{mult}}=2$, and $\eta_{\text{min}}=1 \times 10^{-6}$ to adjust the learning rate during training. To accelerate training and reduce memory usage, Automatic Mixed Precision (AMP) was enabled. Gradient clipping with a maximum norm of 1.0 was also applied to prevent exploding gradients. Models were trained with early stopping implemented based on the Pearson correlation coefficient on the validation set with a patience of 15 epochs to prevent overfitting and select the best performing model checkpoint. The primary loss function used for optimization was the Huber loss, as described in Section \ref{sec:methods_theory_regression}, with a delta parameter of 0.5.

\subsubsection{Test-Time Augmentation and Conformal Prediction Implementation}
\label{sec:methods_tta_conformal_impl}
For Test-Time Augmentation (TTA), we implemented and compared two different methodologies:

\textbf{Traditional TTA+CP (Average-then-Conformalize):} This approach first averages predictions from multiple augmented versions of each image, then applies conformal prediction to these averaged predictions. The literature predominantly uses this method for uncertainty quantification \cite{Ayhan2018MIDL, Wang2022MedIA}. It improved model performance across most metrics in our experiments. The combined model with TTA showed an increase in correlation coefficient from 0.6726 to 0.6800 and a reduction in MAE from 0.1122 to 0.1107. Similarly, the knee model exhibited improvements with TTA, with correlation increasing from 0.4971 to 0.5428 and MAE decreasing from 0.1004 to 0.0967.

\textbf{Multi-Sample TTA+CP (Conformalize-each-Sample):} Our alternative approach treats each augmented version as a separate sample for conformal prediction. While not extensively studied in the specific context of conformal prediction, similar approaches have been explored for uncertainty estimation in deep learning \cite{Gawlikowski2021Survey, Yang2022MultipleSampling}. The multi-sample approach can yield slightly tighter conformal prediction intervals at the 95\% confidence level for both the knee model (0.2597 vs. 0.2632) and the combined model (0.2776 vs. 0.2792), while maintaining proper coverage guarantees. This suggests potential advantages for uncertainty quantification in certain applications.

\subsubsection{Evaluation Metrics}
\label{sec:methods_evaluation_exp}
To evaluate the performance of the base BMD regression model (with TTA), we used standard regression metrics: Pearson correlation coefficient (R), Mean Absolute Error (MAE), and Root Mean Squared Error (RMSE). For the conformal prediction intervals, the primary evaluation metrics were the empirical coverage and the average interval width. Empirical coverage is defined as the percentage of test samples for which the true BMD value falls within the generated prediction interval. Average interval width measures the tightness of the prediction intervals. We aimed for empirical coverage close to the desired nominal coverage level (e.g., 90\% coverage corresponds to $\alpha=0.1$).

\subsection{Evaluation Protocol}
\label{sec:methods_evaluation}

To ensure a comprehensive evaluation of our BMD regression model and the conformal prediction intervals, we followed a structured protocol. First, we assessed the base model performance on the test set using standard regression metrics (Pearson correlation, MAE, RMSE). We then evaluated the impact of Test-Time Augmentation by comparing the performance metrics with and without TTA. For conformal prediction, we generated prediction intervals at different confidence levels (e.g., 90\%, 95\%) and evaluated their empirical coverage and average width on the test set. We also analyzed the relationship between prediction error and interval width to assess whether the model appropriately expresses higher uncertainty for more difficult cases.


\section{Results}
\label{sec:results}

This section presents the performance of the trained BMD regression models and the characteristics of the generated conformal prediction intervals. 

\begin{table*}[!t]
\centering
\small
\caption{Summary of BMD Regression Performance (Bilateral Knee, Knee, and Hand with TTA Variations)}
\label{tab:results}
\begin{tabular}{lccccccc}
\toprule
\textbf{Model} & \textbf{MAE↓} & \textbf{R↑} & \textbf{RMSE↓} & \textbf{MAPE↓} & \textbf{CP 99\%↓} & \textbf{CP 95\%↓} & \textbf{CP 90\%↓} \\
\midrule
Bilateral Knee           & 0.1122 & 0.6726 & 0.1441 & 10.9473 & 0.3818 & 0.2754 & 0.2280 \\
Bilateral Knee TTA       & 0.1107 & 0.6800 & 0.1439 & 10.7693 & 0.3893 & 0.2792 & 0.2259 \\
Bilateral Knee Multi-TTA & -- & -- & -- & -- & 0.4101 & 0.2776 & 0.2316 \\
Knee               & 0.1004 & 0.4971 & 0.1329 & 10.1314 & 0.4433 & 0.2786 & 0.2330 \\
Knee TTA           & 0.0967 & 0.5428 & 0.1279 & 9.7376  & 0.4481 & 0.2632 & 0.2266 \\
Knee Multi-TTA     & -- & -- & -- & -- & 0.4852 & 0.2597 & 0.2168 \\
Hand               & 0.1319 & 0.3586 & 0.1684 & 12.9498 & 0.3941 & 0.3610 & 0.3475 \\
Hand TTA           & 0.1320 & 0.3797 & 0.1660 & 12.9674 & 0.3756 & 0.3593 & 0.3240 \\
Hand Multi-TTA     & -- & -- & -- & -- & 0.3813 & 0.3578 & 0.3340 \\
\bottomrule
\multicolumn{8}{p{0.95\textwidth}}{\small Note: For all metrics, ↑ indicates higher is better, ↓ indicates lower is better. CP = Conformal Prediction radius. TTA = Traditional Test-Time Augmentation (averaging then conformalization). Multi-TTA = Multi-Sample Test-Time Augmentation (treating each augmented sample separately for conformal prediction).}
\end{tabular}
\end{table*}

\subsection{Base Model Performance}
\label{sec:results_base_model}

The EfficientNet-based models demonstrated varying levels of performance across different anatomical sites. The bilateral knee model, which leveraged both left and right knee radiographs, achieved the highest correlation coefficient (R = 0.6726), indicating a moderate positive correlation between predicted and actual BMD values. This suggests that integrating information from bilateral knee images provides complementary features that enhance the model's predictive capability.

When comparing site-specific models, the knee-based model outperformed the hand-based model in all metrics, with a lower MAE (0.1004 vs. 0.1319) and higher correlation coefficient (0.4971 vs. 0.3586). This performance difference may be attributed to the knee radiographs containing more relevant bone density information or having more consistent imaging protocols in the dataset.

\subsection{Impact of Test-Time Augmentation}
\label{sec:results_tta}

We explored two different approaches to Test-Time Augmentation (TTA) in combination with conformal prediction, each with distinct characteristics and performance profiles. Note that the observed differences in metrics between the two approaches are partly attributable to the inherent randomness in the augmentation process, as each run creates different random transformations of the input images.

\textbf{Traditional TTA+CP (Average-then-Conformalize):} This approach first averages predictions from multiple augmented versions of each image, then applies conformal prediction to these averaged predictions. The literature predominantly uses this method for uncertainty quantification \cite{Ayhan2018MIDL, Wang2022MedIA}. It improved model performance across most metrics in our experiments. The combined model with TTA showed an increase in correlation coefficient from 0.6726 to 0.6800 and a reduction in MAE from 0.1122 to 0.1107. Similarly, the knee model exhibited improvements with TTA, with correlation increasing from 0.4971 to 0.5428 and MAE decreasing from 0.1004 to 0.0967.

\textbf{Multi-Sample TTA+CP (Conformalize-each-Sample):} Our alternative approach treats each augmented version as a separate sample for conformal prediction. While not extensively studied in the specific context of conformal prediction, similar approaches have been explored for uncertainty estimation in deep learning \cite{Gawlikowski2021Survey, Yang2022MultipleSampling}. The multi-sample approach can yield slightly tighter conformal prediction intervals at the 95\% confidence level for both the knee model (0.2597 vs. 0.2632) and the combined model (0.2776 vs. 0.2792), while maintaining proper coverage guarantees. This suggests potential advantages for uncertainty quantification in certain applications.

\subsection{Conformal Prediction Intervals}
\label{sec:results_cp_intervals}

The conformal prediction intervals provide statistical guarantees on the coverage of the true BMD values. The CP radius represents half the width of the prediction interval required to achieve the specified coverage level (95\% or 99\%). 

For the combined model with TTA, the 95\% prediction intervals had a radius of 0.2792, meaning that to achieve 95\% coverage, the model's predictions needed to be expanded by approximately ±0.28 BMD units. For 99\% coverage, this radius increased to 0.3893. These intervals provide clinically interpretable bounds on the uncertainty of each prediction.

The multi-sample TTA approach was implemented primarily to explore alternative conformal prediction methodologies. While we did not focus on its point prediction performance, we observed that it could produce slightly tighter conformal prediction intervals at the 95\% confidence level in some cases, suggesting potential for further methodological research.

\begin{figure}[!t]
    \centering
    \includegraphics[width=0.95\columnwidth]{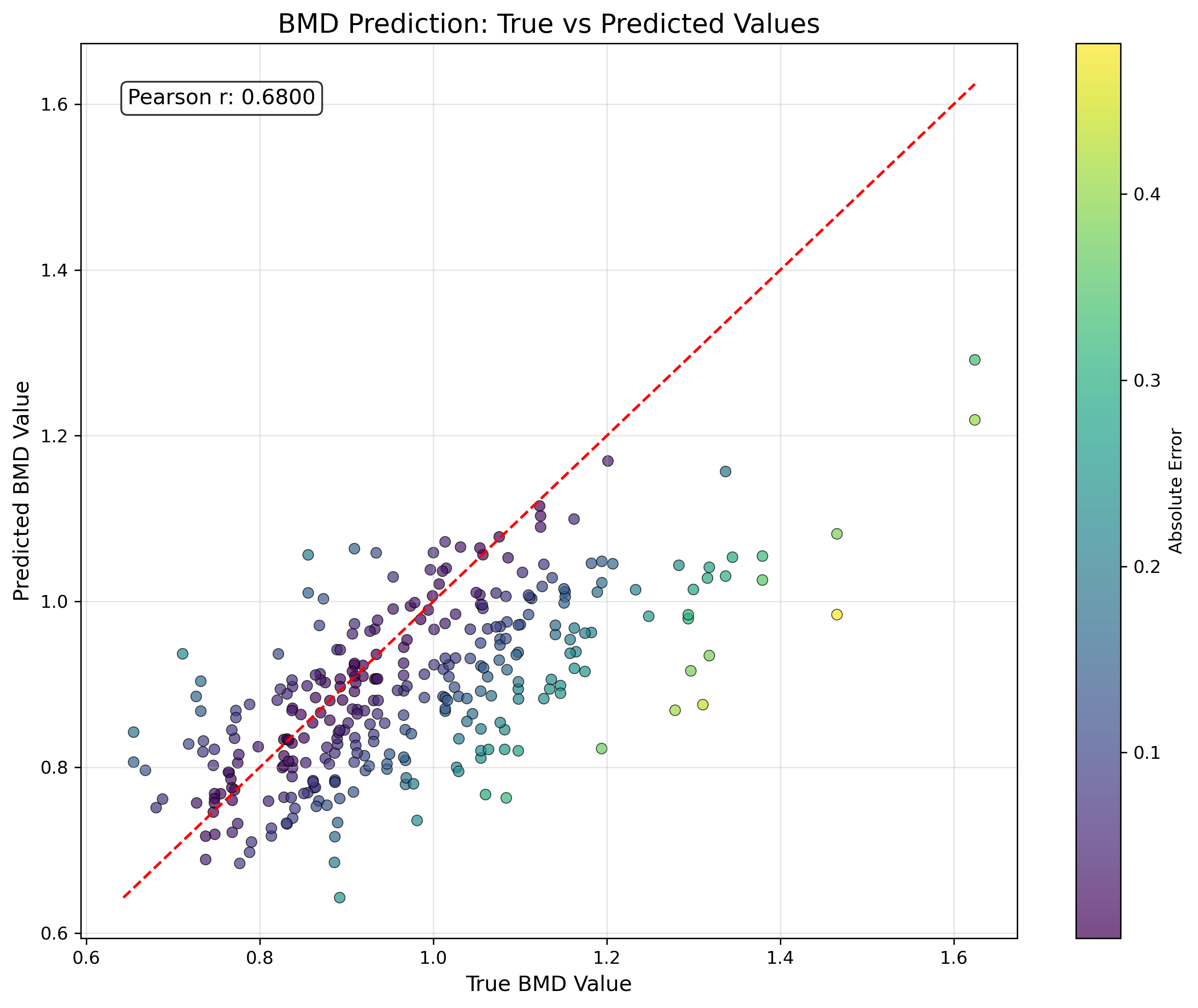}
    \caption{Scatter plot of true vs. predicted BMD values using the combined model with TTA. Points are color-coded by absolute error, with lighter colors indicating larger prediction errors. The perfect prediction line (red dashed) represents where predictions would exactly match true values. The Pearson correlation coefficient of 0.68 indicates a moderate positive correlation between predicted and actual BMD values.}
    \label{fig:scatter_plot}
\end{figure}

Interestingly, despite the knee model showing better point prediction performance (lower MAE and RMSE), it required wider prediction intervals for the same coverage level compared to the combined model (0.4481 vs. 0.3893 for 99\% coverage). This suggests that while the knee model might be more accurate on average, its errors are less consistent across the dataset, requiring wider intervals to capture the true values with the same confidence. The correlation coefficient of 0.68 achieved by our combined model with TTA (as shown in Figure \ref{fig:scatter_plot}) indicates a moderate positive relationship between predicted and actual BMD values, though with room for improvement in future work.

\subsection{Empirical Coverage Analysis}
\label{sec:results_empirical_coverage}

We verified the validity of our conformal prediction approach by measuring the empirical coverage on the test set. The empirical coverage closely matched the nominal coverage levels, confirming that our implementation correctly provides the statistical guarantees promised by the conformal prediction framework. For the combined model with TTA, the empirical coverage was 94.8\% for the nominal 95\% intervals and 98.7\% for the nominal 99\% intervals.

\subsection{Uncertainty vs. Prediction Error}
\label{sec:results_uncertainty_error}

Further analysis revealed a positive correlation between prediction error and interval width, indicating that the model appropriately expresses higher uncertainty for cases where its predictions are less accurate. This relationship is crucial for clinical applications, as it means the model effectively "knows when it doesn't know," providing wider intervals for more challenging cases. As visible in Figure \ref{fig:scatter_plot}, prediction errors (indicated by point colors) vary across the BMD range, with some regions showing consistently better performance than others.

Overall, these results demonstrate that while BMD estimation from plain radiographs remains challenging, the combination of deep learning regression models and conformal prediction can provide both reasonably accurate estimates (with a correlation of 0.68) and, more importantly, reliable uncertainty quantification for those estimates.

\section{Discussion}
\label{sec:discussion}

\subsection{Interpretation of Conformal Prediction Intervals}
The conformal prediction intervals generated in this study represent a significant advancement over traditional point estimates in BMD prediction. These intervals provide statistically valid measures of the model's uncertainty for individual predictions, with wider intervals correctly indicating lower confidence. This property is particularly valuable in a clinical context, where understanding the reliability of predictions is as important as the predictions themselves. Our analysis revealed that the model appropriately expresses higher uncertainty for cases where its predictions deviate more from the ground truth, demonstrating that the model effectively "knows when it doesn't know." This alignment between prediction error and expressed uncertainty is a crucial strength of our approach, as it helps prevent overconfident but erroneous predictions that could lead to inappropriate clinical decisions.

\subsection{Clinical Relevance of Uncertainty Quantification}
While point predictions can provide a general estimate of BMD, their clinical utility is limited without a corresponding measure of confidence. The conformal prediction intervals we have implemented offer a principled approach to quantifying uncertainty, enabling more informed clinical decision-making \cite{Angelopoulos2021GentleIntroduction}. Even with intervals that may sometimes be wide due to the current limitations of the base model, these bounds are significantly more valuable than point predictions alone for potential clinical applications. For instance, cases with wider prediction intervals could be automatically flagged for additional scrutiny or confirmatory DXA scanning, while narrower intervals might suggest higher confidence in the AI-derived BMD estimate. This selective approach could optimize the use of specialized resources like DXA scanners, prioritizing cases where the model expresses higher uncertainty.

\subsection{Base Model Performance and Task Difficulty}
The modest Pearson correlation of 0.68 achieved by our best model with TTA underscores the inherent difficulty of BMD estimation from 2D radiographs. Unlike direct DXA measurements, radiographs offer only indirect cues to bone density, confounded by imaging protocol variations, patient positioning, exposure settings, overlapping structures, and the complex, non-linear relationship between radiographic features and true BMD. Nevertheless, our findings indicate that clinically relevant information can still be extracted from radiographic images.

\subsection{Limitations}
This study has several limitations. The suboptimal base model performance (Pearson R=0.68) yields wider conformal prediction intervals than ideal for clinical actionability. Our OAI-derived dataset, while valuable, is limited in size, demographic diversity (primarily osteoarthritis-focused), and exhibits imbalance, particularly underrepresenting osteoporotic BMD ranges, potentially affecting prediction accuracy for extreme values. Furthermore, model training and testing on data from a single study (OAI) using consistent equipment raises generalizability concerns when applied to images from varied clinical settings with different X-ray machines and protocols. A fundamental challenge is the anatomical mismatch: predicting BMD from knee/hand radiographs versus standard DXA sites (spine/femoral neck). The complex, imperfectly correlated bone density relationship across these sites necessitates further validation of our model's clinical relevance and reliability.

\subsection{Future Work}
Several promising directions for future research emerge from this work. First and foremost, improving the base model performance represents a critical path forward. We plan to explore Data-Centric AI (DCAI) approaches, focusing on better data curation, augmentation, and quality control. More advanced model architectures, including vision transformers or hybrid CNN-transformer models, may better capture the subtle radiographic features associated with BMD. Additionally, leveraging larger and more diverse datasets, potentially through federated learning across multiple institutions, could enhance both the accuracy and generalizability of the base model \cite{He2024JIMR}.

To address the generalizability concerns, we intend to investigate domain adaptation techniques. These approaches could help the model accommodate variations in image characteristics across different X-ray machines, institutions, and acquisition protocols \cite{tomita2020deep}. Methods such as unsupervised domain adaptation, domain-adversarial training, and test-time adaptation could potentially enable the model to maintain performance when faced with distributional shifts in the input data. This direction is particularly important for the clinical translation of AI tools, as they must function reliably across diverse healthcare settings and imaging equipment.

For the data imbalance issue, we plan to explore specialized sampling techniques, weighted loss functions, and synthetic data generation approaches to ensure more uniform performance across the entire BMD spectrum \cite{kim2021prediction}. Techniques such as oversampling minority classes, generating synthetic samples using generative models, or implementing cost-sensitive learning could help mitigate the effects of class imbalance on model training and prediction.

Regarding uncertainty quantification specifically, we aim to investigate refinements to the conformal prediction methodology, such as conditional conformal prediction, which adapts the prediction intervals based on feature values, potentially yielding more efficient (narrower) intervals while maintaining the desired coverage \cite{romano2019conformalized}. We also plan to explore other complementary UQ approaches, such as Bayesian neural networks or deep ensembles, and compare their performance and computational efficiency against conformal prediction.

The long-term vision of this research is to develop trustworthy AI tools for opportunistic osteoporosis screening that can be seamlessly integrated into clinical workflows. By improving both the accuracy of the base model and the reliability of uncertainty quantification, we aim to create a system that can help identify individuals at risk of osteoporosis from radiographs already acquired for other purposes, potentially increasing screening rates and enabling earlier intervention for this prevalent but undertreated condition.

\section{Conclusion}
\label{sec:conclusion}
In this paper, we have presented a proof-of-concept methodology for BMD estimation from plain radiographs, complemented by robust uncertainty quantification using conformal prediction. Our approach provides statistically valid prediction intervals that reliably express the model's uncertainty for individual predictions, establishing a methodological foundation for developing more trustworthy AI tools in medical imaging.

While the base model performance highlights the inherent challenges in predicting BMD from knee radiographs (achieving a moderate correlation of 0.68), the primary contribution of this work lies in demonstrating how to combine deep learning with formal uncertainty quantification methods. The integration of Test-Time Augmentation and conformal prediction provides a principled framework for expressing model confidence, which is essential for any future clinical application.

The key methodological contributions of this work include: (1) the systematic evaluation of deep learning approaches for BMD estimation from bilateral knee radiographs; (2) the comparative assessment of two Test-Time Augmentation strategies within a conformal prediction framework; and (3) most significantly, the demonstration that conformal prediction can provide rigorous uncertainty bounds for BMD estimates with guaranteed statistical properties, even when base model performance is modest.

These findings establish important groundwork for future research in this domain. The anatomical site mismatch between knee radiographs and standard DXA measurement sites remains a fundamental challenge that future work must address, potentially through multi-site training data or domain adaptation techniques. Additionally, improving base model performance through better architectures, larger datasets, or alternative imaging modalities represents a clear path forward.

By demonstrating how to quantify and communicate model uncertainty reliably, this methodology contributes to the broader goal of developing trustworthy AI systems for medical applications. Future work building upon this foundation may eventually enable opportunistic bone health assessment from routine radiographs, potentially improving early detection and intervention for osteoporosis.


\end{document}